\documentclass{article}

\usepackage[final]{neurips_2022}

\usepackage[utf8]{inputenc} % allow utf-8 input
\usepackage[T1]{fontenc}    % use 8-bit T1 fonts
\usepackage{hyperref}       % hyperlinks
\usepackage{url}            % simple URL typesetting
\usepackage{booktabs}       % professional-quality tables
\usepackage{amsfonts}       % blackboard math symbols
\usepackage{nicefrac}       % compact symbols for 1/2, etc.
\usepackage{microtype}      % microtypography
\usepackage{xcolor}         % colors

\usepackage{times}
\usepackage{latexsym}
\usepackage{tabularx}
\usepackage[textsize=scriptsize]{todonotes}
\usepackage{multirow}
\usepackage{amsmath}
\usepackage{mathtools}
\usepackage{amssymb}
\usepackage{amsthm}
\usepackage{tablefootnote}
\usepackage{caption}
\usepackage{subcaption}
\usepackage{bbm}
\usepackage{comment}
\usepackage{xspace}
\usepackage{algorithm}
\usepackage{proof}
\usepackage{stmaryrd}
\usepackage{bm}
\usepackage{pifont}
\usepackage{soul}
\usepackage{wrapfig}

\usepackage{microtype}

\newcommand{\STAB}[1]{\begin{tabular}{@{}c@{}}#1\end{tabular}}

\newcommand{\opt}{\textsc{OPT}}
\newcommand{\dav}{GPT-3}
\newcommand{\davone}{InstructGPT}
\newcommand{\davtwo}{text-davinci-002}
\newcommand{\gpt}{\textsc{GPT-3}}
\newcommand{\synth}{\textsc{Synth}}
\newcommand{\hotpot}{\textsc{AdvHotpot}}
\newcommand{\esnli}{\textsc{E-SNLI}}

\title{The Unreliability of Explanations in Few-shot Prompting for Textual Reasoning}

\author{%
  Xi Ye \quad \quad Greg Durrett \\
  Department of Computer Science\\
  The University of Texas at Austin\\
  \texttt{\{xiye,gdurrett\}@cs.utexas.edu} \\
}

\begin{document}

\maketitle

\begin{abstract}
Does prompting a large language model (LLM) like \gpt{} with explanations improve in-context learning?
We study this question on two NLP tasks that involve reasoning over text, namely question answering and natural language inference.
% For these tasks, we find that including explanations \gpt{}'s prompt and having the model generate them only mildly improves accuracy over standard few-shot learning,\todo{need to update because of divergence between 002 and davinci/OPT} contrary to recent results on symbolic reasoning tasks \citep{scratch,chain}.
We test the performance of four LLMs on three textual reasoning datasets using prompts that include explanations in multiple different styles. For these tasks, we find that including explanations in the prompts for OPT, GPT-3 (davinci), and InstructGPT (text-davinci-001) only yields small to moderate accuracy improvements over standard few-show learning. However, text-davinci-002 is able to benefit more substantially.

We further show that explanations generated by the LLMs may not entail the models' predictions nor be factually grounded in the input, even on simple tasks with extractive explanations. However, these flawed explanations can still be useful as a way to verify LLMs' predictions post-hoc. Through analysis in our three settings, we show that explanations judged by humans to be good---logically consistent with the input and the prediction---more likely cooccur with accurate predictions. Following these observations, we train calibrators using automatically extracted scores that assess the reliability of explanations, allowing us to improve performance post-hoc across all of our datasets.\footnote{Data and code available at \url{https://github.com/xiye17/TextualExplInContext} }
\end{abstract}

\begin{figure}[h]
% \vspace{-0.125in}
\centering
\includegraphics[width=\linewidth,trim=190 260 190 260,clip]{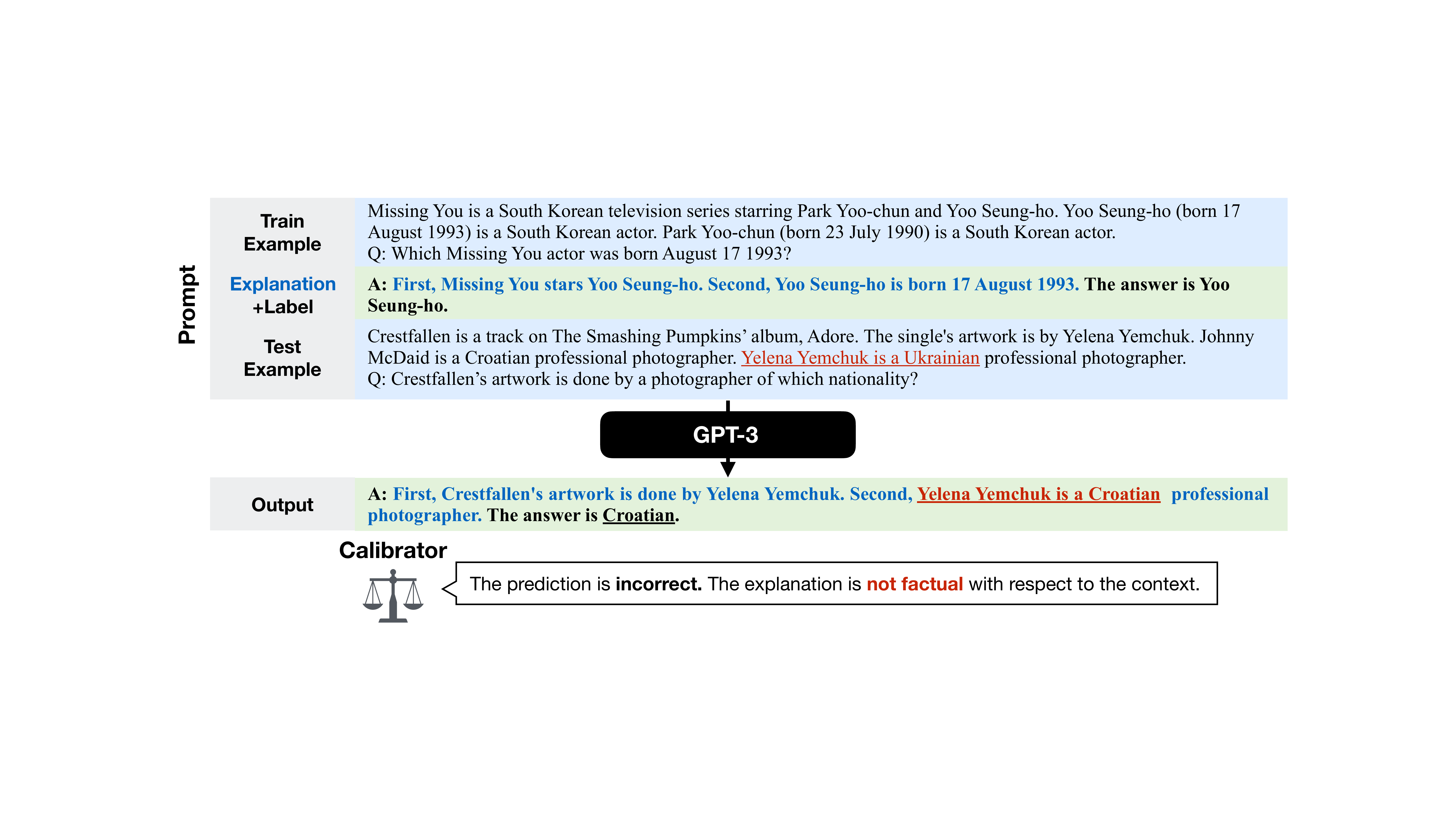}
\caption{Prompting GPT-3 with explanations. By including explanations in the in-context examples, we can cause GPT-3 to generate an explanation for the test example as well. In this case, the generated explanation is nonfactual, despite the simple reasoning involved here. However, we show this nonfactuality actually provides a signal that can help calibrate the model.}
% \vspace{-0.075in}
\label{fig:motivation}
\end{figure}

\section{Introduction}

% Recent\todo{figure} breakthroughs in pretraining have e growing capabilities of large language models in few-shot learning. In particular, in-context learning, i.e., teaching large language models (e.g., GPT3) a new task by prompting a with a few training examples, has shown promising performance across a wide range of tasks. This new paradigm for few-shot learning is particularly exciting because it allows data-efficient and tuning-free deployment of large language models as black boxes to a new task or a new domain.\todo{first paragraph maybe too generic?}

%However, the predictions obtained through in-context learning are often poorly calibrated \citep{calibrate} and lack interpretability. Moreover, past work has pointed out the wayward behavior of in-context learning: it does not faithfully leverage the instructions or even the labels of the few-shot examples provided in the prompt \citep{min2022,webson2021}.

Recent scaling of pre-training has empowered large language models (LLMs) to learn NLP tasks from just a few training examples ``in context,'' without updating the model's parameters \citep{gpt3}. However, this learning process is still poorly understood: models are biased by the order of in-context examples \citep{calibrate} and may not leverage the instructions or even the labels of the examples in the ways one expects \citep{min2022,webson2021}. 
Existing tools for interpreting model predictions have high computational cost
\citep{ribeiro-lime} or require access to gradients \citep{simonyan-interp2013,sundararajan-ig}, making them unsuitable for investigating in-context learning or explaining the predictions of prompted models.

One appealing way to gain more insight into predictions obtained through in-context learning is to let the language model ``explain itself'' \citep{scratch,chain,palm,Marasovi2021,lampinen2022}.
In addition to input-label training pairs in context, one can prompt the language model with an explanation for each pair and trigger the model to generate an explanation for its prediction (Figure~\ref{fig:motivation}).
Prompting with explanations introduces much richer information compared to using labels alone, which might guide the inference process and allow the model to learn more information from the examples.

% show that explanations only mildly improve performance when plugged into the prompt (Figure~\ref{fig:motivation}) across three different datasets spanning QA and NLI.\todo{this paragraph needs updating based on the davinci 000-002 results, also maybe clarification about the scope of textual reasoning we examine}

In this work, we investigate the nature of the explanations that LLMs generate and whether they can improve few-shot in-context learning for textual reasoning tasks, specifically QA and NLI. Recent prior work that finds success with this approach largely targets symbolic reasoning tasks with a very different structure, such as math word problem solving \citep{scratch,chain}. We experiment on three different datasets spanning QA and NLI with four LLMs: OPT, GPT-3 (davinci), InstructGPT (text-davinci-001), and text-davinci-002. The results suggest that explanations only substantially improve accuracy for text-davinci-002, but give a smaller improvement or even hurt the performance with the other LLMs.

% We therefore conduct more fine-grained manual analysis on the quality of the generated explanations to study (1) the capabilities of \gpt{} to generate factual and consistent explanations, even if they explanations do not directly boost the performance. (2) the connection between the reliability of an explanation and the correctness of a prediction.  Specifically, we check whether an explanation for a given input example is correctly grounded in the input (factuality) and whether the explanation correctly entails the final prediction (consistency). Interestingly, as shown in Figure~\ref{fig:motivation}, \gpt{} tends to generate consistent explanations that account for the predictions, but the explanations may not faithfully be grounded in the context in the inputs.
% Furthermore, our analysis suggests a nonfactual explanation more likely indicates a wrong prediction, compared to a factual explanation.

Surprisingly, we find that the explanations generated by LLMs can be \textbf{unreliable}, even for a very simple synthetic dataset.
We evaluate the explanations along two axes: \emph{factuality}, whether the explanation is correctly grounded in the input, and \emph{consistency}, whether the explanation entails the final prediction. LLMs tend to generate consistent explanations that account for the predictions, but the explanations may not be factual, as as shown in Figure~\ref{fig:motivation}.
Furthermore, our analysis suggests an unreliable explanation more likely indicates a wrong prediction compared to a reliable explanation.

% These results are negative ones about the capabilities of \gpt{}, but such a connection sheds light on the possibility of leveraging the quality of explanations to judge the correctness of the predictions post-hoc. Suppose there exists an oracle that can assess the factuality of a explanation, we can then easily use it to rule out incorrect predictions, and hence improve \gpt{}'s in-context learning performance. Unfortunately, such an oracle does not exist, and is itself a challenging problem. However, we can still rely on simple heuristic rules as a proxy of such an oracle to approximate the factuality of an explanation, and use that to rerank or calibrate \gpt{}'s predictions, which successfully yields more accurate or better calibrated predictions across all the datasets.
Despite LLMs' failures here, we can still benefit from model-generated explanations by using them for calibration. If we are able to automatically assess the reliability of an explanation, we can allow an LLM to return a null answer when its explanation is unreliable, since the prediction in this case is less likely to be correct. Unfortunately, there is no automated way to perfectly assess the reliability, but we can extract features that approximately reflect it. We use these features to calibrate InstructGPT's\footnote{Throughout our paper, we primarily test on InstructGPT for two reasons. First, it was the most capable model available at the time we were conducting the majority of our experiments. Second, it still has significant room to improve on the datasets we explore in this work. This setting is a representative testbed for the situation where an LLM-based system does not yet give satisfactory performance on a target task, causing the system designer to turn to explanations in prompts to improve things.} predictions, and successfully improve the in-context learning performance across all the datasets.

% In summary, our main findings are:
% \begin{itemize}
% \item Simply plugging explanations into the prompt does not necessarily improve the in-context learning performance for textual reasoning tasks.
% \item \gpt{} can generate mostly consistent explanations, but these explanations might not be faithfully grounded in the inputs.
% \item The factuality of an explanation can serve as an indicator for the correctness of the corresponding prediction.
% \item Using features that can approximate the factuality of explanations, we successfully use explanations to improve the in-context learning performance across all tasks.
% \end{itemize}

In summary, our main findings are:
(1) Simply plugging explanations into the prompt does not always substantially boost the in-context learning performance for textual reasoning.
(2) LLMs generate explanations consistent with their predictions, but these explanations might not be factually grounded in the inputs.
(3) The factuality of an explanation can serve as an indicator for the correctness of the corresponding prediction.
(4) Using features that can approximate the factuality of explanations, we successfully use explanations to improve the in-context learning performance across all tasks.

\section{Does Prompting with Explanations Improve In-Context Learning?}

In this paper, we specifically focus on tasks involving reasoning over natural language. These are tasks where explanations have been traditionally studied \citep{esnli,rajani2019}, but which are more complex than tasks like sentiment analysis which are well explained by extractive rationales \citep{zaidan2007,eraser}. 
We experiment on two tasks, reading comprehension question answering (QA) and natural language inference (NLI), on three English-language datasets. For each dataset, we create a test set with 250 examples.%\footnote{In our preliminary experiments we found a test set of 250 examples is enough to yield statistically significant differences between the models. Using larger datasets would greatly increase the cost of this work.}

\subsection{Datasets}
\label{sec:datasets}
\paragraph{Synthetic Multi-hop QA (\synth{})} In order to have a controlled setting where we can easily understand whether explanations are factual and consistent with the answer, we create a synthetic multi-hop QA dataset. Shown in Figure~\ref{fig:data_exs}, each example in this dataset asks a bridge question (using the terminology of \cite{hotpot}) over a context consisting of supporting facts paired with controlled distractors. This dataset is carefully designed to avoid spurious correlations, giving us full understanding over the correct reasoning process and the explanation for every example, which naturally consists of the two supporting sentences. See Appendix~\ref{app:synth} for full details of this dataset.\footnote{This dataset is inspired by task 15 of the bAbI dataset \citep{babi}. In our preliminary experiments with some of the other bAbI tasks, we found poor performance from InstructGPT similar to our results on \synth{}, both with and without explanations.}

\begin{figure}[t]
    \centering
    \scriptsize
      \renewcommand{\tabcolsep}{1.0mm}
    \begin{tabular}{l|rl}
    \toprule
         \multirow{4}{*}{\STAB{\rotatebox[origin=c]{90}{\sc Synth}}}& \textbf{Context:} & Christopher agrees with Kevin. Tiffany agrees with Matthew. Mary hangs out with Danielle. James hangs out \\ &&with Thomas. Kevin is a student. Matthew is a plumber. Danielle is a student. Thomas is a plumber.  \\
         &\textbf{Question:}& Who hangs out with a student? \\
        %  & \textbf{Answer:} & Mary \;\;\quad \textbf{Explanation:} Mary hangs out with Danielle. Danielle is a student. \\
         &\textbf{Answer:}& Mary \;\;\quad \textbf{Explanation:} Danielle is a student and Mary hangs out with Danielle. \\
         \midrule
         \multirow{3}{*}{\STAB{\rotatebox[origin=c]{90}{\sc E-SNLI}}}& \textbf{Premise:}& A toddler in a green jersey is being followed by a wheelchair bound woman in a red sweater past a wooden bench. \\
         &\textbf{Hypothesis:}& A toddler is walking near his wheelchair bound grandmother.\\
         &\textbf{Label:}& Neither\quad \textbf{Explanation:} the woman may not be his grandmother.\\
         \bottomrule
    \end{tabular}
    \caption{A \synth{} example and an \esnli{} example. See Figure~\ref{fig:expl_error} for \hotpot{} examples. }
    \vspace{-0.2in}
    \label{fig:data_exs}
\end{figure}

\paragraph{Adversarial HotpotQA (\hotpot{})}  We also test on the English-language Adversarial HotpotQA dataset \citep{hotpot,advhotpot}. We use the adversarially augmented version since InstructGPT achieves high performance on the distractor setting of the original dataset. We make a challenging set of examples by balancing sets of questions on which InstructGPT makes correct and incorrect predictions. The context of each question includes two ground truth supporting paragraphs and two adversarial paragraphs. Full details of preprocessing the \hotpot{} dataset can be found in Appendix~\ref{app:hotpotadv}.

For \hotpot{}, we manually annotated explanations for the training examples. Figure~\ref{fig:motivation} shows an example of such an explanation, highlighted in orange. We could use the supporting sentences as the explanations, but we found they are usually too verbose and not sufficient, e.g., with anaphors that resolve outside of the supporting sentences. Therefore, we manually annotate a set of explanations which clearly describe the reasoning path for each question.

\paragraph{E-SNLI} \esnli{} \citep{esnli} is an English-language classification dataset commonly used to study explanations, released under the MIT license. Shown in Figure~\ref{fig:data_exs}, each example consists of a premise and a hypothesis, and the task is to classify the hypothesis as entailed by, contradicted by, or neutral with respect to the premise. As a notable contrast to the other datasets, the explanations here are more \textit{abstract} natural language written by human annotators, as opposed to mostly constructed from extracted snippets of context.% We use the official version, and randomly sample 250 examples as the test set from the original test split.

\subsection{Baselines}
We study the effectiveness of plugging in explanations by comparing the in-context learning performance of prompting with or without explanations. Prompting without explanations resembles the standard few-shot in-context learning approach (\textbf{Few-Shot}). To incorporate explanations into the prompt, we consider the following two most commonly used paradigms:

\textbf{Explain-then-Predict (E-P)} prepends an explanation before the label (Figure~\ref{fig:motivation}). The language model is expected to generate an explanation first followed by the prediction. The prompting style of past work involving computational traces can be categorized into this paradigm, including \citet{scratch} and \citet{chain}. This approach is also called a pipeline model in other literature on training models using explanations \citep{jacovi2021,wiegreffe2021}.

\textbf{Predict-then-Explain (P-E)} generates the explanation after the prediction. Unlike E-P, the predicted explanation does not influence the predicted label, since we use greedy inference and the explanation comes afterwards. However, the explanations in the prompt still impact the predictions.

\subsection{Setup}

For few-shot learning, we use roughly the maximum allowed shots in the prompt that can fit the length limit of \opt{} \citep{metaopt} and \gpt{} \citep{gpt3}, which is 16 for \synth{}, 6 for \hotpot{}, and 32 for \esnli{}, respectively.\footnote{This contrasts with recent work like \citet{calibrate} that focuses on improving performance in the 1-4-shot setting; by using more data we achieve much stronger results on our tasks.}
We experiment with four LLMs, including \opt{} (175B), \dav{} (davinci), \davone{} (text-davinci-001), and \davtwo{}. % \footnote{We did not experiment with smaller models, as these are much worse at in-context learning \citep{metaopt, gpt3}.} 
\opt{} and \dav{} are trained using the standard causal language modeling objective, whereas \davone{} and \davtwo{} are trained with special instruction data and human annotations. We generate outputs with greedy decoding (temperature set to be 0). Our prompt formats follow those in \citet{gpt3}.
The explanations are inserted before/after the prediction with conjunction words like \emph{because}. Please refer to Appendix~\ref{sec:app_prompt} for full prompts. Because the results of in-context learning vary with the examples presented in the input prompt, for each dataset, we randomly sample multiple groups of training shots, and report the mean and standard deviation of the results (subscript). We use 5 groups for \davone{}, the primary LM we are using throughout our paper, and 3 groups for the rest.

\subsection{Results}
\label{sec:icl_results}

\begin{table}[t]
  \caption{Results of prompting with explanations on four large language models. Using explanations leads to small to moderate improves performance on OPT, GPT-3, and InstructGPT, and has more prominent effects on text-davinci-002.}
  \label{tab:promptingLLMs}
  \centering
  \scriptsize

  \begin{tabular}{clccc}
    \toprule
    
     && \synth{} & \hotpot{} & \esnli{} \\
    \midrule
    \multirow{3}{*}{ OPT (175B)} 
    & \sc Few-Shot & {\bf 40.5}\textsubscript{2.8}  & {49.7}\textsubscript{2.6}  & {\bf 44.0}\textsubscript{3.8}     \\
    \cmidrule{2-5}
    & \sc E-P & { 29.6}\textsubscript{0.5} & {\bf 52.6}\textsubscript{6.5} & 39.3\textsubscript{7.8}      \\
    & \sc P-E & 40.2\textsubscript{2.6} & {43.3}\textsubscript{4.5} & {43.4}\textsubscript{1.6}  \\
    
    \midrule
     \multirow{3}{*}{ GPT-3} 
    & \sc Few-Shot & {49.5}\textsubscript{0.6}  & 49.1\textsubscript{6.2} & 43.3\textsubscript{5.7}     \\
    \cmidrule{2-5}
    & \sc E-P & {47.1}\textsubscript{2.8} & {\bf 54.1}\textsubscript{4.1} & 40.4\textsubscript{4.5}      \\
    & \sc P-E & {\bf 51.3}\textsubscript{1.8} & 48.7\textsubscript{4.6} & {\bf 48.7}\textsubscript{2.4}  \\
    
    \midrule
    \multirow{3}{*}{InstructGPT} 
    & \sc Few-Shot & {54.8}\textsubscript{3.1}  & 53.2\textsubscript{2.3} & 56.8\textsubscript{2.0}     \\
    \cmidrule{2-5}
    & \sc E-P & {\bf 58.5}\textsubscript{2.1} & {\bf 58.2}\textsubscript{4.1} & 41.8\textsubscript{2.5}      \\
    & \sc P-E & 53.6\textsubscript{1.0} & 51.5\textsubscript{2.4} & {\bf 59.4}\textsubscript{1.0}  \\
    \midrule
    \multirow{3}{*}{text-davinci-002} 
    & \sc Few-Shot & {72.0}\textsubscript{1.4}  & 77.7\textsubscript{3.2} & 69.1\textsubscript{2.0}     \\
    \cmidrule{2-5}
    & \sc E-P & {\bf 86.9}\textsubscript{3.8} & {\bf 82.4}\textsubscript{5.1} & {\bf 75.6}\textsubscript{7.6}      \\
    & \sc P-E & 81.1\textsubscript{2.8} & 77.2\textsubscript{4.8} & { 69.4}\textsubscript{5.0}  \\
    \bottomrule
  \end{tabular}
  \vspace{-0.15in}
\end{table}

As shown in Table~\ref{tab:promptingLLMs},  \opt{}, \dav{}, and \davone{} show small to moderate improvements from using explanations for textual reasoning tasks. On the two QA tasks, \synth{} and \hotpot{}, {\sc E-P} improves the performance of \davone{}, the best among these three LMs, from 54.8 to 58.5 and 56.8 to 59.4, respectively.\footnote{For \synth{}, we also tried using an alternative style of explanations (reversing the order of the two sentences in the explanations), which leads to mild performance degradation.} On \esnli{}, {\sc P-E} outperforms {\sc Few-Shot} by 2.6, whereas {\sc E-P} substantially lags {\sc Few-Shot}. Comparing {\sc E-P} against {\sc P-E} on \synth{} and \esnli{}, {\sc E-P} typically degrades performance (except on \synth{} for \davone{}) and P-E is inconsistent across the different models, whereas {\sc E-P} consistently leads to performance improvements on \hotpot{}. There is no single winner between the two paradigms of using explanations; choosing the most effective way is task-specific. Overall, vanilla LLMs (\opt{} and \dav{}) see limited benefit from producing explanations, and even the Instruct-series \davone{} does not see substantial improvements.

The only exception is \davtwo{}. \davtwo{} greatly benefits from explanations in the prompt across all three tasks, and {\sc E-P} is consistently more effective than {\sc P-E}. However, it is unclear what contributes to this difference. As far as we are aware, the differences between \davtwo{} and \davone{} are not described in any publication or blog post.\footnote{One publicly-described difference is the addition of editing and insertion, discussed at \url{https://openai.com/blog/gpt-3-edit-insert/}, but this does not explain the performance differences we observe.} Comparing \dav{} and \davone{}, we see the move to Instruct series models is \emph{not} sufficient to explain the difference. Given the lack of transparency with this model, we hesitate to make scientific claims about the results it yields. \looseness=-1

% One possibility is that 002 is an updated version of 001 that includes more Instruct data collected using the API. One hypothesis for the improvement is data leakage from our test set. Because we started running experiments for this work in late 2021, it is conceivable that text-davinci-002 was trained on human-written completions for our data. Another hypothesis is that text-davinci-002 features T0-like fine-tuning on some available datasets such as HotpotQA, which would also change the interpretation of the results.

Our results do not suggest immediate strong improvements from incorporating explanations across all LLMs, even for our synthetic dataset, contradicting recent prior work. This can be attributed to the difference between the tasks we study. The tasks that receive significant benefits from using explanations in \citet{scratch} and \citet{chain} are all program-like (e.g., integer addition and program execution), whereas the tasks in this work emphasize textual reasoning grounded in provided inputs. In fact, in \citet{chain} and \citet{palm}, explanations only show mild benefit on open-domain QA tasks like StrategyQA \citep{geva2021} that are closer to our setting.

\begin{figure}[t]
    \centering
    \scriptsize
    \begin{tabular}{l|l}
    \toprule
   \multirow{7}{*}{\STAB{\rotatebox[origin=c]{90}{\bf Nonfactual}}} &  Pedro Rubens! The individual chapters were published into 64 "tankōbon" by Kodansha.\\
 & Yōko Shōji (born 4 June 1950, in Mobara, Chiba) is a Japanese manga artist. She is best known for writing "Seito Shokun! \\ 
& Mulder Scully! The individual chapters were published into 14 "tankōbon" by Kodansha. \\
& \textcolor{red}{Seito Shokun! The individual chapters were published into 24 "tankōbon" by Kodansha between.} \\
& Q: How many chapters does Yōko Shōji's most famous manga have? \\
& A: First, Yōko Shōji's most famous manga is "Seito Shokun!". \textcolor{red}{ Second, "Seito Shokun!" has 64 chapters.}  The answer is \textcolor{red}{64}. \\
    \midrule
    \multirow{7}{*}{\STAB{\rotatebox[origin=c]{90}{\bf Inconsistent}}} & Tim Minchin (December 29, 1808 July 31, 1875) was the President of the United States.\\
    &  Andrew Johnson (December 29, 1808 July 31, 1875) was the President of the United States.\\
    & George Andrew Atzerodt (June 12, 1835 – July 7, 1865) was a conspirator, with John Wilkes Booth. \\
    & Jesse Andrew Williams (June 12, 1835 – July 7, 1865) was a conspirator, with John Wilkes Booth. \\
    & Q: Who was older, George Atzerodt or Andrew Johnson? \\
    & A: First, George Atzerodt was born on \textcolor{red}{June 12, 1835}. Second, Andrew Johnson was born on \textcolor{red}{December 29, 1808}. The \\
    &answer is \textcolor{red}{George Atzerodt}.  \\
    \bottomrule
    \end{tabular}
    \caption{Explanations generated for \hotpot{}. InstructGPT may generate nonfactual explanations containing hallucination (red) or inconsistent explanations contradicting the answer (red). }
    \label{fig:expl_error}
      \vspace{-0.15in}
\end{figure}

\section{Can LLMs Generate Factual and Consistent Explanations? }
\label{sec:quality}

\begin{table}[t]
    \caption{Left: factuality (Fac) and consistency (Con) of the generated explanations. Right: the \% of the examples whose explanation factuality/consistency is congruent with the prediction accuracy. In general, LLMs tend to generate consistent but less likely factual explanations.}
  \label{tab:qualityLLMs}
  \centering
    \scriptsize   
    \begin{tabular}{clccccc}
    \toprule
  && \bf Acc & \bf Fac & \bf Con &  \bf Acc=Fac & \bf Acc=Con\\
    \midrule
      & \multicolumn{6}{c}{\textit{reliability of explanations generated by \davone{}}} \\
      \cmidrule{2-7}
\multirow{5}{*}{\davone{}} & 
{\sc Synth} ({\sc E-P})&  58.4 & 72.8  & 64.8  & 66.5 & 68.8\\
    & {\sc Synth} ({\sc P-E})&  54.8 & 51.6  & 95.2  & \bf 89.6 & 57.2\\
    \cmidrule{2-2}
    &{\sc AdvHP} ({\sc E-P}) &  62.0  & 79.6 & 91.2   & \bf 80.0 &  68.4  \\
    &{\sc AdvHP} ({\sc P-E}) &  54.0  & 69.2 & 82.0   & \bf 77.6 & 67.2  \\
    \cmidrule{2-2}
    &\esnli{} ({\sc P-E})   &62.0& $-$ & 98.8 & $-$ & 62.0\\
        \midrule
         &  \multicolumn{6}{c}{\textit{reliability of explanations generated by other LLMs on \synth{}}} \\
    \cmidrule{2-7}

        \multirow{2}{*}{\sc OPT (175B)} &
    {\sc Synth} ({\sc E-P})&  30.0 & 77.2 & 47.2 & 45.6 & 58.8\\
    & {\sc Synth} ({\sc P-E})&  39.6 & 64.0 & 81.2 & \bf 69.2 & 49.6\\
    \cmidrule{1-1}
    \multirow{2}{*}{\dav{}} & 
        {\sc Synth} ({\sc E-P})& 46.8 & 59.2 & 64.8 & \bf 66.8 & 61.2\\
    & {\sc Synth} ({\sc P-E})&  52.4 & 52.4 & 83.2 & \bf 78.4 & 58.0\\
    \cmidrule{1-1}
    \multirow{2}{*}{\davtwo{}} & 
    {\sc Synth} ({\sc E-P})&  86.0 & 91.6 & 85.2 & \bf 91.2 & 84.8\\
    & {\sc Synth} ({\sc P-E})& 81.6 & 83.2 & 96.4 & \bf 95.8 & 82.8\\
    \bottomrule
  \end{tabular}
        \vspace{-0.15in}

\end{table}

Prompting LLMs with explanations and having models generate them may not guarantee higher performance on our tasks. But what about the quality of the model-generated explanations themselves? We assess the reliability of the explanations for the three datasets, measured in terms of two aspects.

\textbf{Factuality} refers to whether a generated explanation is faithfully grounded in the corresponding input context (context for QA and premise/hypothesis pair for NLI). A factual explanation should not contain hallucinations that contradict the context. See Figure~\ref{fig:expl_error} for a nonfactual explanation. %In some ways, soundness also connects to \emph{plausibility}, i.e., whether the explanation makes sense for the human based on the given context. Here we use term soundness which is more closely tied with the settings in our work.

\textbf{Consistency} measures if the explanation entails the prediction. Our concept of consistency resembles plausibility as described in \citet{jacovi2021}, in that we assess whether the prediction follows from the explanation \textbf{as perceived by a human}. See Figure~\ref{fig:expl_error} for an inconsistent explanation.

\begin{wrapfigure}{r}{0.40\textwidth}
    \centering
    \includegraphics[width=0.40\textwidth,trim=30 10 30 0,clip]{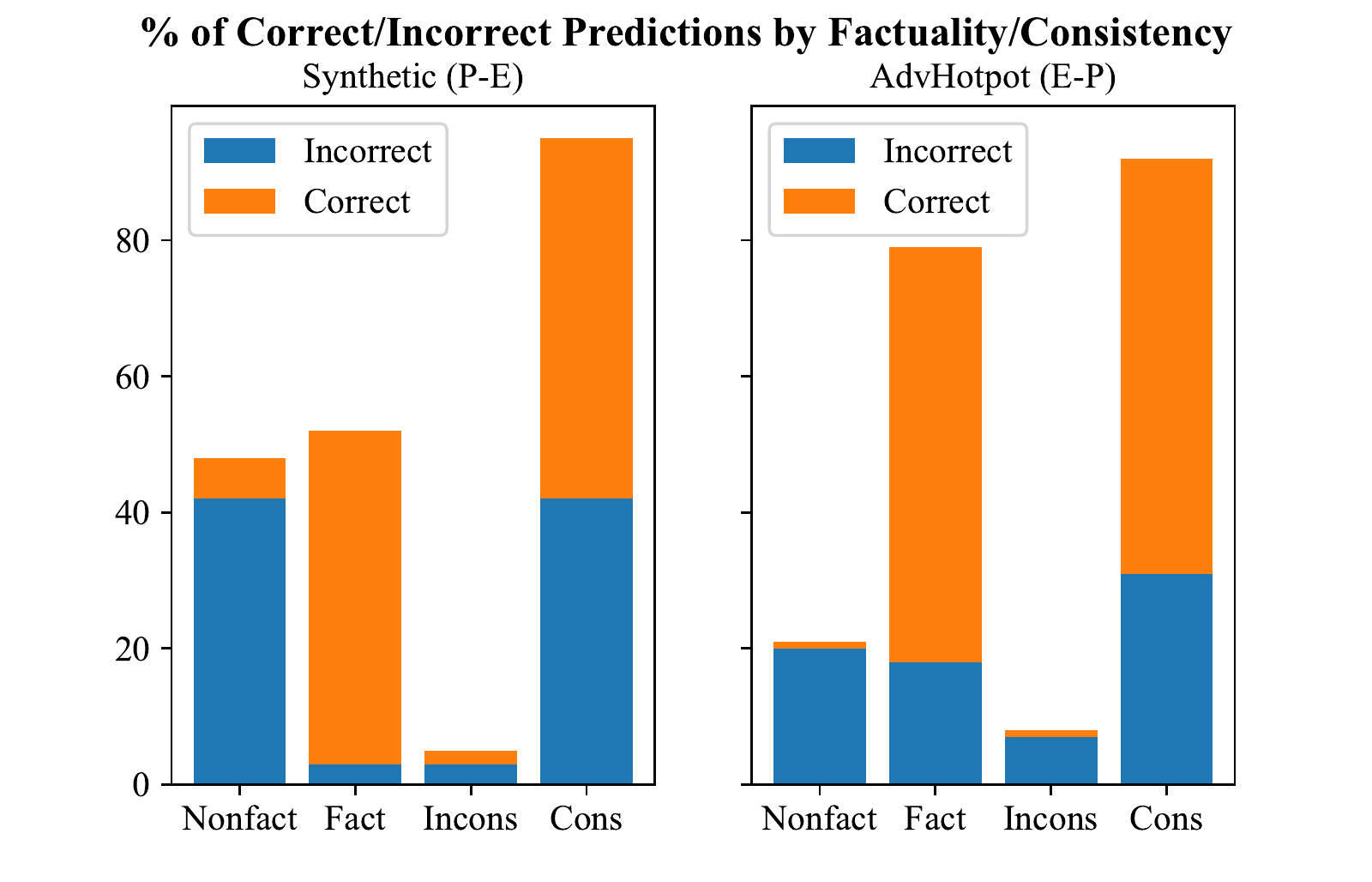}
    \caption{Explanations are more likely to be nonfactual than to be inconsistent, and a nonfactual explanation usually indicates an incorrect prediction.}
    \vspace{-0.20in}
    \label{fig:barchart}   
\end{wrapfigure}

For \synth{}, we use rules to automatically judge whether an explanation is factual and consistent on all four LLMs. For \hotpot{} and \esnli{}, the authors manually inspected the explanations generated by \davone{} and annotated them for these two characteristics (more details in Appendix~\ref{app:annotation}). Note for each setting, the results are based on the explanations and predictions obtained with a single set of training shots. We only show the results of {\sc P-E} on \esnli{}, as {\sc E-P} is substantially worse here. % For each dataset, we annotate the explanations generated by whichever is the best paradigm between {\sc E-P} and {\sc P-E}. We have also assessed the reliability for {\sc E-P} on \synth{} and {\sc P-E} on \hotpot{} in Appendix~\ref{app:add_quality}.

% \begin{figure}
%     \begin{minipage}{0.57\textwidth}
%     \captionsetup{type=table}
%     \caption{Left: factuality (Fac) and consistency (Con) of the generated explanations. Right: the \% of the examples whose explanation factuality/consistency is congruent with the prediction accuracy. In general, \gpt{} tends to generate consistent but less likely factual explanations.}
%   \label{tab:quality}
%   \centering
%   \scriptsize
%   \renewcommand{\tabcolsep}{1.0mm}
%   \begin{tabular}{lccc|cc}
%     \toprule
%     % \multicolumn{2}{c}{Part}                   \\
%     % \cmidrule(r){1-2}
%   & Acc & Fac & Con &  Acc=Fac & Acc=Con\\
%     \midrule
%     {\sc Synth} ({\sc E-P})&  58.4 & 72.8  & 64.8  &66.5 & 68.8\\
%     {\sc Synth} ({\sc P-E})&  54.8 & 51.6  & 95.2  &89.6 & 57.2\\
%     \cmidrule{1-1}
%     {\sc AdvHP} ({\sc E-P}) &  62.0  & 79.6 & 91.2   & 80.0 &  68.4  \\
%     {\sc AdvHP} ({\sc P-E}) &  54.0  & 69.2 & 82.0   & 77.6 & 67.2  \\
%     \cmidrule{1-1}
%     \esnli{} ({\sc P-E})   &62.0& $-$ & 98.8 & $-$ & 62.0\\
%     \bottomrule
%   \end{tabular}
%     \end{minipage}
%     \begin{minipage}{.43\textwidth}
%     \centering
% \includegraphics[width=0.9\textwidth,trim=20 0 20 0,clip]{figures/group_v3.pdf}
% \caption{Nonfactual explanations usually indicate an incorrect prediction.}
% \label{fig:barchart}
%   \vspace{-0.15in}

%     \end{minipage}
% \end{figure}

% \begin{figure}[t]

% \end{figure}

\paragraph{Results}
We summarize the results in Table~\ref{tab:qualityLLMs}. We only report consistency on \esnli{}, as the explanations for \esnli{} often require some external commonsense knowledge which cannot be easily grounded in the inputs or judged as true or false (examples in Appendix~\ref{sec:app_add_exs}). The results suggest a disconnect between the model predictions and the ``reasoning'' in explanations.
On \davone{}, though using explanations improves its performance across three tasks, the generated explanations are \textit{unreliable} (upper section), even for the straightforward synthetic setting. Comparing the factuality of explanations for \synth{} generated by \dav{}, \davone{}, and \davtwo{}, we see that instruction tuning improves the factuality, but even the most powerful \davtwo{} still fails to generate explanations that are perfectly grounded in the input context.
Overall, LLMs tend to generate consistent explanations (>80\% for all three datasets with the right prompt structure), but the explanations are less likely to be factual, which is concerning as they can deceive a user of the system into believing the model's answer.

\subsection{Reliability of Explanations and Prediction Accuracy}
\label{sec:correlation}
LLMs may hallucinate problematic explanations, but this could actually be advantageous if it gives us a way of spotting when the model's ``reasoning'' has failed. We investigate the connection between the reliability of an explanation and the accuracy of a prediction and ask whether a reliable explanation indicates an accurate prediction. (This resembles the linguistic calibration of \citet{Mielke2020LinguisticCT}, but using a different signal for calibration.)

As shown in Table~\ref{tab:qualityLLMs} (right), accuracy and factuality/consistency are typically correlated, especially factuality. By knowing whether an explanation is factual, we can guess the model's accuracy a high fraction of the time (Accuracy = Factuality). A nonfactual explanation very likely means an incorrect prediction on the \synth{} dataset across all four LLMs. On \hotpot{}, factuality and \davone{}'s prediction correspond 80.0\% of the time, substantially surpassing the prediction accuracy itself. We show fractions of correct and incorrect predictions when the explanations are factual/nonfactual and consistent/inconsistent in Figure~\ref{fig:barchart} for two of our settings. Factual explanations are much more likely paired with correct predictions compared to nonfactual explanations. Consistency is also connected to accuracy but is an inferior indicator compared to factuality in general (Table~\ref{tab:qualityLLMs}). \looseness=-1

% \section{Using Heuristics as a Proxy to Improve In-Context Learning Performance}
\section{Calibrating In-Context Learning using Explanations}
From Section~\ref{sec:correlation}, we see that a human oracle assessment of the factuality of an explanation could be of substantial use for calibrating the corresponding prediction. Can we automate this process?

We first show how to achieve this goal on the perfectly controlled \synth{} dataset (Section~\ref{sec:calib_synth}). On our other two datasets, we use surface lexical matching to approximate semantic matching and give real-valued scores approximately reflecting factuality. Following past work on supervised calibration \citep{selectiveqa,chendurrett2021,interpcalib}, we can learn a calibrator that tunes the probabilities of a prediction based on the score of its explanation (Section~\ref{sec:calib_framework}). We show such a calibrator can be trained with a handful of examples beyond those used for in-context learning and successfully improve the in-context learning performance on realistic datasets.\footnote{This procedure does require extra data. However, it provides a natural avenue for using a small number of additional examples that otherwise would be \emph{impossible} to incorporate into this procedure, when the size of the context actually limits the amount of data for in-context learning.} We note that, as mentioned before, the experiments in this section are conducted on \davone{}.

\subsection{Motivating Example: Improving \synth{} Dataset}
\label{sec:calib_synth}
We first show how post-hoc calibration functions in the controlled \synth{} setting, where we can simply check the factuality of an explanation. Since the generated explanation always follows the format ``\texttt{\small B is [profession] and A [verb] B.}'' (example in Figure~\ref{fig:data_exs}), we can split the explanation into two sentences. The explanation is factual if and only if each of the two sentences exactly matches one of the sentences in the context.

We use the assessment to improve the performance of {\sc P-E} for \synth{}, where a nonfactual explanation typically indicates an incorrect prediction. This gives us a way to reject presumably incorrect answers. Specifically, we iterate through the top 5 candidate answers (restricted by the API) given by \davone{} and reject any answer-explanation pair if the explanation is nonfactual until we find a factual one. This procedure dramatically improves the accuracy from 52.4\% to 74.8\%. Note that this \synth{} dataset is a challenging task given its lack of reasoning shortcuts: for reference, neither {\sc RoBERTa} \citep{roberta} nor {\sc DeBERTa} \citep{deberta} finetuned with 16 examples can achieve an accuracy surpassing 50\%. With the help of the explanations and the checking procedure, we can use \davone{} to achieve strong results using few-shot learning.

\subsection{Learning-based Calibration Framework}
\label{sec:calib_framework}

%Unfortunately, there does not exist such an oracle on realistic datasets, but we can still use some simple heuristics to approximately decide the factuality, and use this to improve in-context learning performance. 

\paragraph{Framework}
% Before giving details of the heuristics
We now introduce the framework that can leverage the factuality assessment of an explanation to calibrate a prediction. Let $\boldsymbol{p}$ be the vector of predicted probabilities associated with each class label in NLI (or the probability score of predicted answer in QA). Let $v$ be a scalar value extracted from the explanation to describe the factuality. Then, we can adjust the probabilities accordingly using a linear model: $\boldsymbol{\hat{p}} = \mathrm{softmax}(W[\boldsymbol{p};v] + b),$
% \vspace{-0.1in}
% \footnotesize
% $$
% \normalsize
% \noindent
where $\boldsymbol{\hat{p}}$ is the tuned probabilities.

Our calibration framework is extended from classical calibration methods \citep{platt1999,guocalib,calibrate}, which apply an affine transformation on the probabilities alone: $\boldsymbol{\hat{p}} = \mathrm{softmax}(W\boldsymbol{p} + b)$. In contrast, we use an additional factor $v$ in calibration to incorporate the factuality assessment of the explanation.

There are a small number of parameters ($W$ and $b$) that need to be trained in such a calibration framework. %We will rely on a few more examples in addition to the examples we use in the prompt to train the calibrator. Note there is \textbf{no} requirement for explanation annotations for the additional examples. We only need to generate the explanations for them using annotated examples, and extract factors to construct training data points used to train the calibrator.
We will rely on a few more examples in addition to the shots we use in the prompt to train the calibrator. Specifically, we use the prompt examples to generate the predictions and explanations for these extra examples, and extract predicted probabilities, factors, and target probabilities triples to construct training data points used to train the calibrator.
Note this procedure requires \textbf{no} explanation annotations for the extra examples.

\paragraph{Approximating Factuality}

We approximate the factuality using lexical overlap between the explanations and the inputs, which we found to work fairly well for our tasks.

{\bf \hotpot{}:} We use an explanation consisting of two sentences (examples in Figure~\ref{fig:expl_error}) as an illustration. Let $\mathcal{E}=(E^{(1)},E^{(2)})$ be the generated explanation, where $E^{(1)}$ and $E^{(2)}$ are the two sentences, and the $E^{(i)}=(e_1,e_2,\cdots)$ contain tokens $e_1,e_2,\cdots$. Similarly, let $\mathcal{P}=(P^{(1)},P^{(2)},P^{(3)},P^{(4)})$ be the context paragraphs, and $P^{(i)}=(p_1,p_2,\cdots)$ be the tokens. The factuality estimation of one explanation sentence $E^{(i)}$ is defined as: $ \mathcal{V}(E^{(i)})=\max_{P\in\mathcal{P}} \frac{|E^{(i)} \cap P| }{|E^{(i)}|}$.

% \vspace{-0.1in}
% \footnotesize
% \begin{equation*}
%     \mathcal{V}(E^{(i)})=\max_{P\in\mathcal{P}} \frac{|E^{(i)} \cap P| }{|E^{(i)}|}.
% \end{equation*}
% \normalsize

Intuitively, the factuality score for a sentence $E$ is defined as the maximum number of overlapping tokens over all paragraphs $P$, normalized by the number of tokens in $E$. We then define the factuality score for the whole explanation as $\mathcal{V}(\mathcal{E})=\min_{E\in\mathcal{E}}\mathcal{V}(E)$, as it requires all sentences to be factual in order to make the entire explanation factual.\footnote{Alternatively, one might use a fine-tuned NLI model as a proxy \citep{chendurrett2021}. However, our focus is on the pure black-box setting, and we avoid models that require substantial amounts of data to make work.}

{\bf \esnli{}:} The explanations of \esnli{} do not really involve a concept of factuality. Nevertheless, we use an analogous score following the same principle by viewing the premise as the context. Let $E=(e_1,e_2,\cdots)$ be the explanation and  $P=(p_1,p_2,\cdots)$ be the premise. We simply score the explanation by  $\mathcal{V}(E)=\frac{|E| \cap |P| }{|E|}.$ The more an explanation overlaps with the premise, the more factual we judge it to be.\looseness=-1

\subsection{Calibrating \esnli{}}

\begin{wraptable}{r}{0.5\textwidth}
    \centering
        \renewcommand{\tabcolsep}{0.7mm}
    \vspace{-0.2in}
    \caption{Accuracy (mean$_\textrm{std dev}$) of various methods on \esnli{} under different data conditions. \textbf{L} denotes number of labels (as well as the total number of examples); \textbf{E} denotes the number of explanations. Calibrating using explanations successfully improves the performance of in-context learning. }
    \scriptsize
    \label{tab:exp_snli}
    \begin{tabular}{lcccc}
    \toprule
   \bf w/o Explanation & \bf 32L & \bf 64L & \bf 96L & \bf 128L \\
    \midrule
    RoBERTa & 40.1\textsubscript{4.7} & 43.0\textsubscript{5.1} & 49.0\textsubscript{5.2} & 54.9\textsubscript{4.8}  \\
    \cmidrule(r){1-1}
    \sc Few-Shot & 56.8\textsubscript{2.0}& $-$ & $-$ & $-$ \\
    \sc Few-Shot(NN)  & $-$ & $-$ & $-$ & 58.9\textsubscript{1.0} \\
    \sc Few-Shot+ProbCal & 61.9\textsubscript{3.8} & 62.4\textsubscript{2.6} & 63.2\textsubscript{2.9} & 63.9\textsubscript{1.2}\\
    \midrule
    % \multicolumn{5}{c}{with Explanation} \\
    \bf w/ Explanation    & \bf32L+32E & \bf 64L+32E &  \bf96L+32E & \bf128L+32E \\
     \midrule
    % E-P & 51.2 & $-$ & $-$ & $-$ \\
    \sc P-E & 59.4\textsubscript{2.0} & $-$ & $-$ & $-$ \\
    \cmidrule(r){1-1}
    \sc P-E+ProbCal & \bf 64.4\textsubscript{1.8} & 65.4\textsubscript{1.2} & 65.4\textsubscript{1.6} & 65.4\textsubscript{1.9} \\
    \sc P-E+ExplCal & 64.2\textsubscript{2.6} & \bf 65.8\textsubscript{1.3} & \bf 67.6\textsubscript{1.6} & \bf 68.5\textsubscript{1.2} \\
    \cmidrule(r){1-1}
    \sc P-E+Zhang & 63.0\textsubscript{3.2} & 65.2\textsubscript{2.2} & 65.4\textsubscript{1.5} & 65.9\textsubscript{2.5} \\
    \bottomrule
    \end{tabular}
        \vspace{-0.1in}
\end{wraptable}

\paragraph{Setup} For E-SNLI, we use calibration methods to postprocess the final probabilities. Unlike classical temperature scaling \citep{platt1999}, note that the methods we use here can actually change the prediction; we will therefore evaluate on \emph{accuracy} of the calibrated model.

We study the effectiveness of our explanation-based calibrator under different training data sizes varying from 32 to 128. Recall that we only require explanation annotations for 32 data points, and only need the labels for the rest to train the calibrator. For \esnli{}, we calibrate {\sc P-E}, which is shown to be more effective than {\sc E-P} in this setting (Section~\ref{sec:icl_results}).

% As in \citet{calibrate}, large language models can be poorly calibrated when used for in-context learning. 
\paragraph{Baselines} We provide the performance of fine-tuned \textsc{RoBERTa} \citep{roberta} model as a reference, finding this to work better than DeBERTa \citep{deberta}. To isolate the effectiveness of using explanations for calibration, we introduce three additional baselines using non-explanation-based calibrators. We apply the probability-based calibrator as described in Section~\ref{sec:calib_framework} on the results obtained on few-shot learning ({\sc Few-Shot+ProbCal}) and predict-then-explain pipeline ({\sc P-E+ProbCal}). We note that the parameters of these calibrators are trained using the additional data points, as opposed to being heuristically determined as in \citet{calibrate}. Furthermore, we experiment with a recently proposed supervised calibrator from \citet{zhangeunsol}, which uses the CLS representations from an additional language model as features in the calibrator. The probabilities are tuned using $\boldsymbol{\hat{p}} = \mathrm{softmax}(W[\boldsymbol{p};\boldsymbol{h}] + b)$, where $\boldsymbol{h}$ is the CLS representation. Since we do not have access to the embeddings obtained by \gpt{}, we use {\sc RoBERTa} to extract the vectors instead. We use such a calibrator on top of our best-performing base model, {\sc P-E}, resulting {\sc P-E+ \citet{zhangeunsol}}.

% The calibrator is essentially the same liner transformation as in the Explanation-Based reranker except for excluding the signal $e$, i.e., $\hat{p} = \mathrm{softmax}(Wp + b).$

% This re-ranker resembles the calibrator in \citet{calibrate}, but we train the parameters rather than heuristically determining them with de-contextualized prompt, and hence our variant is expected to be more accurate.

Limited by the maximum prompt length, in-context learning is not able to take as input the additional data used for training the calibrator. For a fair comparison, we can allow the in-context model to use this data by varying the prompts across test examples, dynamically choosing the prompt examples to maximize performance. Choosing closer data points for prompting is a common and effective way of scaling up the training data size for in-context learning \citep{shin2021,Liu2021WhatMG}. Following \citet{Liu2021WhatMG}, we test the performance of choosing nearest neighbors for the prompt based on CLS embedding produced by a {\sc RoBERTa} model \citep{roberta}, referred as {\sc Few-Shot(NN)}. It is worth clarifying that the \textsc{Few-Shot} and {\sc Few-Shot+ProbCal} approaches use the same set of 32 training shots in the prompt for every test example, whereas the shot sets vary from example to example in \textsc{Few-Shot(NN)}.

%Lastly, w\footnote{We also tried a more recent {\sc DeBERTa} model , but found it to be worse than {\sc RoBERTa} in the few-shot \esnli{} setting. }  

\paragraph{Results}

We show the results in Table~\ref{tab:exp_snli}.
%We vary the number of training data from 32 to 128. Note that for the approaches requiring explanations (the bottom section in Table~\ref{tab:esnli}), explanations are only needed for 32 examples while only labels are required for the rest. Hence there is only considerable annotation overhead.
We use 5 different groups of training examples and report the mean and standard deviation across the groups.
% We note that for Few-Shot, E-P, and P-E, we only report the results using 32 examples (limited by the maximum length of prompt).
For \textsc{Few-Shot(NN)}, we only report the results obtained using 128 examples; results using a smaller number of examples will be worse than this.

Under 128 training examples, applying a trained calibrator on top of prompting with explanation (i.e., \textsc{P-E+ExplCal}) achieves the best accuracy of 68.5\%, which is 12\% higher than the performance of the vanilla uncalibrated few-shot in-context learning (\textsc{Few-Shot}). \textsc{P-E+ExplCal} also outperforms \textsc{Few-Shot+ProbCal} and \textsc{P-E+ProbCal} by 5\% and 3\%, respectively. Using explanations is more effective than using probabilities alone. In addition, \textsc{P-E+ExplCal} also outperforms  {\sc P-E+\citet{zhangeunsol}}, whose performance is on par with \textsc{P-E+ProbCal}. This suggests the additional CLS information is not very helpful in this setting.
% Besides, calibrating using explanations produces more stable results than, demonstrated by the differences between the standard deviation of \textsc{Few-Shot+ProbCalib}) and \textsc{P-Shot+ProbCalib}, especially with fewer training examples (32-96).

As the data size increases from 32 to 128, the performance of the explanation-based calibrator keeps improving notably, whereas the performance of probability-based calibrators nearly saturates at a data size of 96. The performance of \textsc{Few-Shot(NN)} with 128 training instances only improves the performance by 3.3\%, compared to \textsc{Few-Shot} with 32 training instances. Choosing nearest neighbors as the shots, while being effective when having access to a large amount of data, is not helpful in the extreme data-scarce regime. Calibrating using explanations is an effective way of using a few extra data points that cannot fit in the prompt, which is a pitfall of standard in-context learning.

Finally, \textsc{RoBERTa} finetuned using 128 shots only achieves an accuracy of 54.9\%, lagging the performance of \gpt{} based models. The limited training data size is insufficient for finetuning smaller language models like {\sc RoBERTa}, but is sufficient for \textsc{P-E+ExplCal} to be effective.

% \paragraph{Setup \& Baseline}
% Do we need to delete this paper if BERT gets > 70 with 128 samples.

% My impression is that the results of strict few-shot (especially 32, 64, 96) could be very random, it might be somewhat true (but less serious) with 128 as well.
% \begin{enumerate}
%     \item Reranking seems to be more random/sensitive w.r.t. training data than non-rereanking
%     \item Of cause, more groups of 128 training sample
%     \item More Testing data? now it's 250, maybe we need 500?
%     \item For each 128 training group, the cost is like \$125
% \end{enumerate}

\subsection{Calibrating \hotpot{}}
\label{sec:calib_hotpot}
% \paragraph{SelectiveQA Setting} For the \hotpot{} dataset, we only tune the confidence scores of the predicted answer as oppose to rejecting answers in \synth{}, since we do not have an approach that can precisely reject invalid answers. Nevertheless, better calibrated confidence scores are still valuable  as we can selectively answer a question, i.e., the SelectiveQA setting \citep{selectiveqa}. Instead of giving a wrong answer, the model can opt to abstain when it does not have a confident answer. Therefore, a better calibrated model can give correct answers to more question while avoiding making mistakes compared to a poorly calibrated model. More intuitively, if a calibrator can opt to choose the top 50\% of questions from a dataset that it is most confidant to answer, the performance on the 50\% of examples selected by a well calibrated model will be higher than those of a poorly calibrated model. We use the Area Under Coverage-Accuracy Curve (AUC) to evaluate how well a model is calibrated as in past literature \citep{selectiveqa,chendurrett2021,zhangeunsol, garg2021, interpcalib}. Examples of the curves can be seen in Figure~\ref{fig:auc_curves}.

\begin{figure}
    \begin{minipage}{0.49\textwidth}
    \centering
     \captionsetup{type=table}
    \caption{AUC scores (mean$_\textrm{std dev}$) on \hotpot{} under different data conditions. \textbf{L} and  \textbf{E}  denotes the number of label annotations and explanation annotations, respectively. Explanation-based calibration successfully improves the performance on top of prompting with explanations.}
    \label{tab:exp_hotpot}
    \scriptsize
    \begin{tabular}{lccc}
    \toprule
    % \multicolumn{4}{c}{without Explanation} \\
\bf   w/o Explanation &\bf 6L & \bf 32L & \bf 64L  \\
  \midrule
   \sc Few-Shot & 59.6\textsubscript{2.4} & $-$ & $-$ \\
\sc Few-Shot(NN)& $-$ & $-$ & 61.3\textsubscript{0.9} \\
    \midrule
    % \multicolumn{4}{c}{with Explanation} \\
 \bf     w/ Explanation &\bf 6L+6E & \bf 32L+6E  & \bf64L+6E \\
    % \midrule
    \midrule
    \sc E-P & {\bf 64.4}\textsubscript{2.9} & $-$ & $-$ \\
  \cmidrule(r){1-1}
%   \sc  P-E+ExplCalib &  $-$ &  62.2$\pm$2.8 & 62.6$\pm$3.1\\
  \sc  E-P+ExplCal &  $-$ &  {\bf 66.0}\textsubscript{3.9} & {\bf 68.8}\textsubscript{3.0} \\
    % P-E + Conf Rerank & {\color{red} 70.4} & 68.0 & 66.0 & 66.4 \\
    % P-E + Expl Rerank & 58.8 & 69.2 & 71.2 & 73.2 \\
\cmidrule(r){1-1}
  \sc  E-P+Zhang &  $-$ &  65.6\textsubscript{3.9} & 66.1\textsubscript{3.2}\\
    \bottomrule
    \end{tabular}
    % \vspace{-0.125in}
    \end{minipage}
    \hspace{0.1in}
    \begin{minipage}{0.45\textwidth}
    \centering
    \includegraphics[width=0.95\linewidth,trim=0 0 0 0,clip]{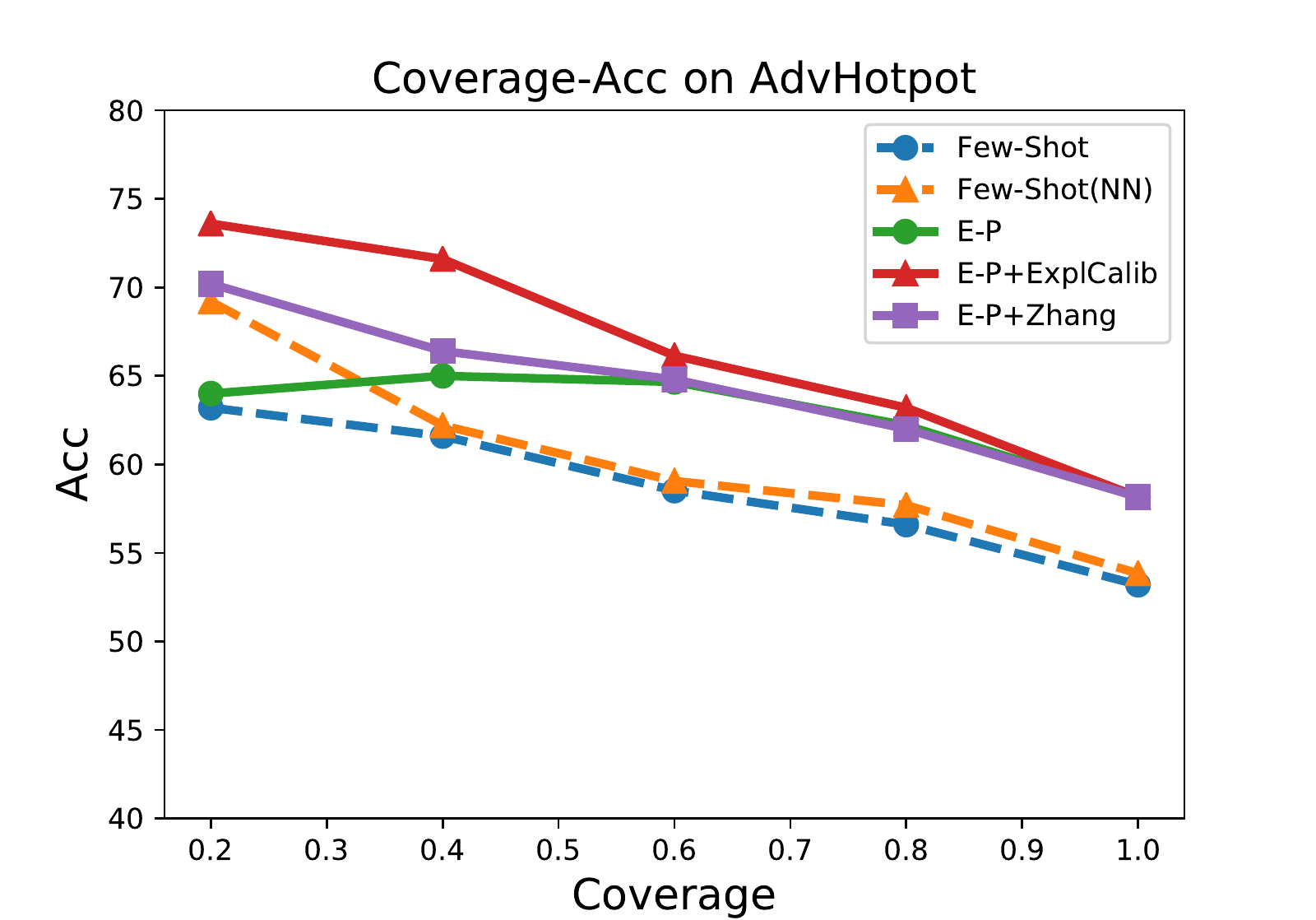}
    \caption{Coverage-Acc curves of various methods on \hotpot{}. {\sc E-P+ExplCal} is better calibrated compared to uncalbrated {\sc E-P} as well as the other approaches.}
        % \vspace{-0.125in}
    \label{fig:auc_curves}
    \end{minipage}
\end{figure}

\paragraph{Setup} For the \hotpot{} dataset, our calibration takes the form of tuning the confidence scores of the predicted answers to better align them with the correctness of predictions. These confidence scores can be used in a ``selective QA'' setting \citep{selectiveqa}, where the model can abstain on a certain fraction of questions where it assigns low confidence to its answers. We use the \emph{area under coverage-accuracy curve} (AUC) to evaluate how well a model is calibrated as in past literature \citep{selectiveqa,chendurrett2021,zhangeunsol, garg2021, interpcalib}. The curve plots the average accuracy with varying fractions (coverage) of questions being answered (examples in Figure~\ref{fig:auc_curves}). For any given coverage, a better calibrated model should be able to identify questions that it performs best on, hence resulting a higher AUC.
% More intuitively, if a calibrator can opt to choose the top 50\% of questions from a dataset that it is most confidant to answer, the performance on the 50\% of examples selected by a well calibrated model will be higher than those of a poorly calibrated model.

We experiment with training data set sizes of 6, 32, and 64. We report the results averaged from 5 trials using different training sets. For \hotpot{}, we calibrate {\sc E-P}, which is shown to be more effective than {\sc P-E} in this setting (Section~\ref{sec:icl_results}). Our approach is also effective for calibrating {\sc P-E}; please refer to Appendix~\ref{app:add_hotpot} for details.

% \paragraph{Calibration Method}
% We tune the confidence scores on AdvHotpot using based on the soundness values computed as above as well as the original confidence scores. Let $p$ be the original confidence score, and $e$ be the factor computed using the explanation, still, we apply a linear transformation to tune the confidence $\hat{p} = \mathrm{softmax}(W[p;e])$. There are esentially only two parameters in $W$ to optimize, which can be done using only handful of data points.

\paragraph{Results}

We show the AUC scores in Table~\ref{tab:exp_hotpot}. By leveraging explanations, {\sc E-P+ExplCal} successfully achieves an AUC of 68.8, surpassing both {\sc Few-Shot} by 7 points and {\sc E-P} by 4 points. We note this is a substantial improvement, given that the upperbound of AUC is constrained by the accuracy of the answers and cannot reach 100. Figure~\ref{fig:auc_curves} shows the coverage-accuracy curves of various methods averaged across the 5 training runs. {\sc E-P+ExplCal} always achieves a higher accuracy than its uncalibrated counterpart, {\sc E-P}, under a certain coverage, and the gap is especially large in the most confident intervals (coverage < 50\%). {\sc E-P+\citet{zhangeunsol}} is able to calibrate the predictions on this dataset, but still lags our explanation-based calibrator, {\sc E-P+ExplCal}.

In addition, the explanation-based calibrator can be effective with as few as 32 examples. This is because there are only two parameters (the probability of predicted answer and the explanation-based factor) in the calibrator, which can be easily learned in this few-shot setting. Comparing {\sc E-P+ExplCal} against {\sc Few-Shot(NN)}, using nearest neighbors in the prompt is also able to improve the performance compared to using a fixed set of shots ({\sc Few-Shot}), yet our lightweight calibrator can better utilize such a small amount of data, and learn to distinguish more accurate predictions based on the explanations.

\section{Related Work}

% Our investigation is centered around in-context learning \citep{gpt3}, which has garnered increasing interest since the breakthrough of various large pretrained language models. Recent work has been devoted to studying different aspects of in-context learning, including its wayward behaviors \citep{min2022,webson2021} and approaches to overcome them \citep{calibrate}. Our exploration is anchored around the angle of using explanations, which has also been discussed concurrently \citep{scratch,chain,palm}. Our work goes beyond using the explanation in the prompt and further investigates the quality of the explanations generated by \gpt{}, which are found to be unreliable.

Our investigation is centered around in-context learning \citep{gpt3}, which has garnered increasing interest since the breakthrough of various large pretrained language models. Recent work has been devoted to studying different aspects of in-context learning, including its wayward behaviors \citep{min2022,webson2021} and approaches to overcome them \citep{calibrate}, whereas our exploration focuses on using explanations.

\looseness=-1
The utility of explanations for few-shot in-context learning has also been discussed concurrently \citep{scratch,chain, Marasovi2021,palm,lampinen2022,wiegreffe2021reframing}, especially in symbolic reasoning tasks. We differ in that we study more free-form explanations in tasks (QA and NLI, specifically) focusing on textual reasoning over provided contexts. Furthermore, our work focuses on the nature of the explanations generated by LLMs, which are found to be unreliable. Regarding our use of calibration, similar ideas of explanation-based performance estimation have been applied to other tasks \citep{rajani-stacking,evalexplqa,interpcalib}, but we rely on the free-text explanations generated by the model instead of interpretations obtained through post-hoc interpretation techniques.

More broadly, how to use explanations in various forms (textual explanation, highlights, etc.) to train better models is a longstanding problem \citep{zaidan2007}.
% Various techniques have been proposed for this purpose.
Past work has built a series of pipeline models that first generate the explanations and then make predictions purely based on the generated explanations \citep{wiegreffe2021,zhoutan2021,howardchen2022}. Prior research has also explored using explanations as additional supervision to train joint models \citep{hancock2018,dua2020,qed,stacey2022}.
% that jointly predict the labels and explanations \citep{hancock2018,dua2020,qed,stacey2022}.
Another line of work seeks to align the reasoning process of a trained model with the explanations, which is typically done by interpreting a prediction post-hoc through explanation techniques and optimizing the distance between the obtained explanation and ground truth explanation \citep{liuavci2019,rieger20a,plumb2020,erion2021improving,yao2021refining}. These aforementioned methods all update the model parameters and typically require a considerable amount of explanation annotations to be effective. By contrast, our setting treats language models as pure black boxes and only requires few-shot explanations.

% \section{Limitations}

% \greg{should we have a separate section here and say something about PaLM? And cite Bender + Koller and say that explanations are fundamentally ungrounded in some sense?}
    
\section{Discussion \& Conclusion}
\label{sec:conclusion}
% \section{Caveats and Risks of Explanations from Large Language Models}
\paragraph{Caveats and Risks of Explanations from Large Language Models}
Our analysis suggests that LLMs' internal ``reasoning'' does not always align with explanations that it generates, as shown by our consistency results. More concerning, the explanations might not be factually grounded in the provided prompt. This shortcoming should caution against any deployment of this technology in practice: because the explanations are grammatical English and look very convincing, they may deceive users into believing the system's responses even when those responses are incorrect. Section 6 of \citet{bender-stochastic} discusses these risks in additional detail. The fact that language models can hallucinate explanations is also found in other work \citep{zhoutan2021}. This result is unsurprising in some sense: without sufficient supervision or grounding, language models do not learn meaning as distinct from form \citep{bender2020}, so we should not expect their explanations to be strongly grounded. % of meaning This is congruent with the findings of other recent analysis on the issues of in-context learning: it may not faithfully ground its predictions purely on the input prompt \citep{webson2021,min2022}.
%Nevertheless, future work can possibly explore special pre-training techniques for making large language models more closely rely on the explanations or the prompt, more generally.

We have shown that even explanations which don't lead to accuracy gains can still be useful for calibration. However, the lexical overlap feature we use here is a weak signal of explanation correctness (see the example in Figure~\ref{fig:motivation}). Strong enough entailment models should theoretically be able to perform this task and work across a range of tasks without fine-tuning. This explanation assessment model can even be a language model itself trained for this particular propose to approach the verification tasks for a given domain by in-context learning.

% \section{Conclusion}
\paragraph{Conclusion}
We have explored the capabilities of LLMs in using explanations in in-context learning for textual reasoning. Through our experiments with four LLMs and on two QA datasets and an NLI dataset, we find that simply including explanations in the prompt does not always improve the performance of in-context learning. Our manual analysis demonstrates that LLMs tend to generate nonfactual explanations when making wrong predictions, which can be a useful leverage to assess the correctness of the predictions. Lastly, we showcase how to use explanations to build lightweight calibrators, which successfully improve \davone{}'s in-context learning performance across all three datasets.

% \section*{Broader Impacts}\todo{fix title}

% Explanations are tricky because they can be unfaithful...Can potentially mislead users into thinking that models understand when they don't. [cite Cynthia Rudin 2019 Stop Interpreting, Lipton Mythos]

% We believe our stuff can lead to useful things in the future but we actually view this as a warning that these explanations aren't ready for production...

\section*{Acknowledgments}
We would like to thank Eunsol Choi, Ruiqi Zhong, Jocelyn Chen, Zayne Sprague, and Jiacheng Xu
for their helpful feedback on drafts of this work, as well as the
anonymous reviewers for their thoughtful reviews. This work was partially supported by NSF Grant
IIS-1814522, NSF CAREER Award IIS-2145280, a grant from Open Philanthropy, a gift from Salesforce Inc., and a gift from Adobe.

\bibliography{neurips_2022}
\bibliographystyle{neurips_2022}

%%%%%%%%%%%%%%%%%%%%%%%%%%%%%%%%%%%%%%%%%%%%%%%%%%%%%%%%%%%%
\section*{Checklist}

%%% BEGIN INSTRUCTIONS %%%
%The checklist follows the references.  Please
%read the checklist guidelines carefully for information on how to answer these
%questions.  For each question, change the default \answerTODO{} to \answerYes{},
%\answerNo{}, or \answerNA{}.  You are strongly encouraged to include a {\bf
%justification to your answer}, either by referencing the appropriate section of
%your paper or providing a brief inline description.  For example:
%\begin{itemize}
%  \item Did you include the license to the code and datasets? \answerYes{See Section~\ref{gen_inst}.}
%  \item Did you include the license to the code and datasets? \answerNo{The code and the data are proprietary.}
%  \item Did you include the license to the code and datasets? \answerNA{}
%\end{itemize}
%Please do not modify the questions and only use the provided macros for your
%answers.  Note that the Checklist section does not count towards the page
%limit.  In your paper, please delete this instructions block and only keep the
%Checklist section heading above along with the questions/answers below.
%%% END INSTRUCTIONS %%%

\begin{enumerate}

\item For all authors...
\begin{enumerate}
  \item Do the main claims made in the abstract and introduction accurately reflect the paper's contributions and scope?
    \answerYes{}
  \item Did you describe the limitations of your work?
    \answerYes{See Section~\ref{sec:conclusion}.}
  \item Did you discuss any potential negative societal impacts of your work?
    \answerYes{See Section~\ref{sec:conclusion}.}
  \item Have you read the ethics review guidelines and ensured that your paper conforms to them?
    \answerYes{}
\end{enumerate}

\item If you are including theoretical results...
\begin{enumerate}
  \item Did you state the full set of assumptions of all theoretical results?
    \answerNA{}
        \item Did you include complete proofs of all theoretical results?
    \answerNA{}
\end{enumerate}

\item If you ran experiments...
\begin{enumerate}
  \item Did you include the code, data, and instructions needed to reproduce the main experimental results (either in the supplemental material or as a URL)?
    % \answerTODO{}
    \answerYes{}
  \item Did you specify all the training details (e.g., data splits, hyperparameters, how they were chosen)?
     \answerYes{}
        \item Did you report error bars (e.g., with respect to the random seed after running experiments multiple times)?
     \answerYes{}
        \item Did you include the total amount of compute and the type of resources used (e.g., type of GPUs, internal cluster, or cloud provider)?
     \answerYes{We use the \gpt{} Instruct-series API (text-davinci-001).}
\end{enumerate}

\item If you are using existing assets (e.g., code, data, models) or curating/releasing new assets...
\begin{enumerate}
  \item If your work uses existing assets, did you cite the creators?
    \answerYes{See reference \citep{advhotpot} and \citep{esnli}.}
  \item Did you mention the license of the assets?
\answerYes{See Section~\ref{sec:datasets}.}
  \item Did you include any new assets either in the supplemental material or as a URL?
    \answerYes{We included the Synthetic dataset in the supplementary material.}
  \item Did you discuss whether and how consent was obtained from people whose data you're using/curating?
    \answerNo{}
  \item Did you discuss whether the data you are using/curating contains personally identifiable information or offensive content?
    \answerNo{}
\end{enumerate}

\item If you used crowdsourcing or conducted research with human subjects...
\begin{enumerate}
  \item Did you include the full text of instructions given to participants and screenshots, if applicable?
    \answerNA{}
  \item Did you describe any potential participant risks, with links to Institutional Review Board (IRB) approvals, if applicable?
    \answerNA{}
  \item Did you include the estimated hourly wage paid to participants and the total amount spent on participant compensation?
    \answerNA{}
\end{enumerate}

\end{enumerate}

%%%%%%%%%%%%%%%%%%%%%%%%%%%%%%%%%%%%%%%%%%%%%%%%%%%%%%%%%%%%

\newpage

\appendix

\section{Details of Prompts}
\label{sec:app_prompt}
We show examples of the prompts used for \synth{}, \hotpot{}, and \esnli{} in Figure~\ref{fig:synth_prompt}, Figure~\ref{fig:hotpot_prompt}, and Figure~\ref{fig:esnli_prompt}, respectively. Our prompts follow the original formats in \citet{gpt3}. For approaches that use explanations ({\sc E-P} and {\sc P-E}), we insert explanations before/after with necessary conjunction words.

\begin{figure}[h]
    \centering
    \footnotesize
    \begin{tabular}{l}
    \toprule
    \multicolumn{1}{c}{\sc Synthetic: Few-Shot}  \\
    \midrule
     Christopher agrees with Kevin. Tiffany agrees with Matthew. Mary hangs out with Danielle. James hangs out \\ with Thomas. Kevin is a student. Matthew is a plumber. Danielle is a student. Thomas is a plumber.  \\
         Q: Who hangs out with a student? \\
         A: Mary\\
    \midrule
    \multicolumn{1}{c}{\sc Synthetic: E-P}  \\
    \midrule
    Christopher agrees with Kevin. Tiffany agrees with Matthew. Mary hangs out with Danielle. James hangs out \\ with Thomas. Kevin is a student. Matthew is a plumber. Danielle is a student. Thomas is a plumber.  \\
         Q: Who hangs out with a student? \\
         A: Because Danielle is a student and Mary hangs out with Danielle, the answer is Mary.\\
    \midrule
    \multicolumn{1}{c}{\sc Synthetic: P-E}  \\
    \midrule
        Christopher agrees with Kevin. Tiffany agrees with Matthew. Mary hangs out with Danielle. James hangs out \\ with Thomas. Kevin is a student. Matthew is a plumber. Danielle is a student. Thomas is a plumber.  \\
         Q: Who hangs out with a student? \\
         A: Mary, because Danielle is a student and Mary hangs out with Danielle .\\
    \bottomrule
    \end{tabular}
    \caption{Examples of prompts for \synth{}.}
    \label{fig:synth_prompt}
\end{figure}

\begin{figure}[h]
    \centering
    \footnotesize
    \begin{tabular}{l}

    \toprule
    \multicolumn{1}{c}{\sc AdvHotpot: Few-Shot}  \\
        \midrule
        Sir Luigi Arthur Pirandello (12 August 1895 – 4 October 1952) was an John journalist. \\
 Sir Keith Arthur Murdoch (12 August 1885 – 4 October 1952) was an Australian journalist. \\
 Australian Associated Press (AAP) is an Australian news agency. The organisation was established in \\ 1935 by Keith Murdoch.\\
Sir Nikolai Arthur Trubetzkoy (12 August 1896 – 4 October 1952) was an Covington journalist.\\
Q: Australian Associated Press was established by a journalist born in which year?\\
A: 1885 \\
\midrule
    \multicolumn{1}{c}{\sc AdvHotpot: E-P}  \\
        \midrule
        Sir Luigi Arthur Pirandello (12 August 1895 – 4 October 1952) was an John journalist. \\
 Sir Keith Arthur Murdoch (12 August 1885 – 4 October 1952) was an Australian journalist. \\
 Australian Associated Press (AAP) is an Australian news agency. The organisation was established in \\ 1935 by Keith Murdoch.\\
Sir Nikolai Arthur Trubetzkoy (12 August 1896 – 4 October 1952) was an Covington journalist.\\
Q: Australian Associated Press was established by a journalist born in which year?\\
A: First, Australian Associated Press was established by Keith Murdoch in 1935. Second, Keith Murdoch was \\ born in 1885. The answer is 1885. \\
\midrule
   \multicolumn{1}{c}{\sc AdvHotpot: P-E}  \\
        \midrule
        Sir Luigi Arthur Pirandello (12 August 1895 – 4 October 1952) was an John journalist. \\
 Sir Keith Arthur Murdoch (12 August 1885 – 4 October 1952) was an Australian journalist. \\
 Australian Associated Press (AAP) is an Australian news agency. The organisation was established in \\ 1935 by Keith Murdoch.\\
Sir Nikolai Arthur Trubetzkoy (12 August 1896 – 4 October 1952) was an Covington journalist.\\
Q: Australian Associated Press was established by a journalist born in which year?\\
A: 1885. The reasons are as follows. First, Australian Associated Press was established by Keith Murdoch  \\ in 1935. Second, Keith Murdochwas born in 1885. The answer is 1885. \\
\bottomrule
    \end{tabular}
    \caption{Examples of prompts for \hotpot{}.}
    \label{fig:hotpot_prompt}
\end{figure}

\begin{figure}[h]
    \centering
    \footnotesize
    \begin{tabular}{l}
\toprule
 \multicolumn{1}{c}{\sc E-SNLI: Few-Shot}  \\
    \midrule
    A person in black tries to knock the last pin down in a game of bowling.\\
Q: The person is a girl. True, False, or Neither?\\
A: Neither\\
\midrule
    \multicolumn{1}{c}{\sc E-SNLI: E-P}  \\
    \midrule
    A person in black tries to knock the last pin down in a game of bowling.\\
Q: The person is a girl. True, False, or Neither?\\
A: Because not every person is a girl, this answer is Neither.\\

    \midrule
    \multicolumn{1}{c}{\sc  E-SNLI: P-E}  \\
    \midrule
    A person in black tries to knock the last pin down in a game of bowling.\\
Q: The person is a girl. True, False, or Neither?\\
A: Neither, because not every person is a girl.\\

\bottomrule
    \end{tabular}
    \caption{Examples of prompts for \esnli{}.}
    \label{fig:esnli_prompt}
\end{figure}

\section{Details of the \synth{} Dataset}
\label{app:synth}

We create a controlled synthetic multi-hop QA dataset. Each context consists of four reasoning chains, where each chain contains two sentences following a template: ``\texttt{\small A [verb] B. B is [profession].}''. We fill in \texttt{\small A} and \texttt{\small B} in the reasoning chain templates using randomly selected names from a pool of 50 names. To fill in the \texttt{\small  [verb]} and \texttt{\small [profession]} in the four reasoning chain templates, we first select two verbs from a pool of 30 verbs and two professions from a pool of 30 professions. Next, we fill in the four chains using the combination of these two verbs and professions, which give a set of completely symmetric chains. Finally, we sample one reasoning chain from all of the four to derive a asking:  ``\texttt{\small Who [verb] [profession]?}'' (example in Figure~\ref{fig:data_exs}).

Such a design ensures there are no reasoning shortcuts \citep{designchoice}, making it a difficult dataset even despite the regular structure of the task. A {\sc RoBERTa} model needs roughly 500 data points to tackle this problem and achieve near 100\% accuracy on the test set.

\section{Details of the \hotpot{} Dataset}
\label{app:hotpotadv}
We preprocess the original Adversarial HotpotQA dataset \citep{hotpot,advhotpot} in a few ways. We reduce the context length to make it better fit the purpose of testing in-context learning. We use two ground truth supporting paragraphs joined with two adversarial paragraphs to construct the context for each question, instead of using all eight distractors. In addition, we simplify each paragraph by only keeping relevant sentences needed for answering the question (or distracting the prediction); otherwise, the prompt length limit only allows 2-3 examples fit in the input prompt.

We make a challenging test test set of 250 examples by balancing the mix of examples on which prompted \gpt{} makes correct and incorrect predictions. This is done by first running few-shot inference over 1000 examples, and then randomly sampling 125 examples with correct and incorrect predictions, respectively.

Since assessing the accuracy of an answer in QA is hard, and F1 scores do not correlate with the true quality of the answers (e.g., ``United States'' is a correct answer but has 0 F1 score with respect to the provided ground truth answer ``US'') \citep{bulian2022}, we manually assess the correctness of the answers. We observed a high inter-annotator agreement (Cohen's Kappa of 0.84) between the correctness annotations of 100 examples on which the annotations of the authors intersected. Please refer to the supplementary material for these annotations.

This dataset is licensed under the MIT license.
% \section{Additional Experiments on \synth{}}
% \label{app:altsynth}
% placeholder

\section{Details of Reliability Annotations}
\label{app:annotation}
The authors manually inspected the predictions and explanations generated for the 250 \hotpot{} test examples using a single set of training shots, and annotated them for factuality and consistency. We observed a Cohen's Kappa of 0.85 between the factuality annotations of 100 examples (obtained using the {\sc E-P} paradigm) on which the annotations of the authors overlapped.

% \begin{table}[t]
%   \caption{Left: factuality and consistency of the generated explanations across different datasets. Right: the \% of the examples whose explanation factuality is congruent with the prediction accuracy.}
%   \label{tab:app_quality}
%   \centering
%   \begin{tabular}{lccc|c}
%     \toprule
%     % \multicolumn{2}{c}{Part}                   \\
%     % \cmidrule(r){1-2}
%   & Accuracy & Factuality & Consistency & Accuracy = Factuality \\
%     \midrule
%     \synth{} ({\sc E-P})&  50.8 & 52.4  & 100.  &98.8 \\
%     \synth{} ({\sc P-E})&  51.2 & 52.8  & 100.  &98.4 \\
%     \midrule
%     \hotpot{} ({\sc E-P}) &  62.0  & 79.6 & 91.2   & 80.0    \\
%     \hotpot{} ({\sc P-E}) &  54.0  & 69.2 & 82.0   & 77.6    \\

%     \bottomrule
%   \end{tabular}
% \end{table}

% \section{Reliability Evaluation of {\sc E-P} on \synth{} and {\sc P-E} on \hotpot{} }
% \label{app:add_quality}

% In addition to the experiments in Section~\ref{sec:quality}, we evaluate the reliability of  {\sc E-P} on \synth{} and {\sc P-E} on \hotpot{} and present the results in Table~\ref{tab:app_quality}. The observations here follow those in Table~\ref{tab:quality}. The model-generated explanations are more likely to be consistent than to be factual, no matter how the training explanations are being plugged in the prompt.

\section{Calibrating {\sc P-E} on \hotpot{}}
\label{app:add_hotpot}

\begin{table}[h]
    \centering
    \caption{AUC scores of various methods on \hotpot{} under different data conditions. Explanations are also effective for calibrating {\sc P-E}.}
    \label{tab:app_hotpot}
    \begin{tabular}{lccc}
    \toprule
    % \multicolumn{4}{c}{without Explanation} \\
\bf   w/o Explanation &\bf 6L & \bf 32L & \bf 64L  \\
  \midrule
   \sc Few-Shot & {\bf 59.6}\textsubscript{2.4} & $-$ & $-$ \\
\sc Few-Shot(NN)& $-$ & $-$ & 61.3\textsubscript{0.9} \\
    \midrule
    % \multicolumn{4}{c}{with Explanation} \\
 \bf     w/ Explanation &\bf 6L+6E & \bf 32L+6E  & \bf64L+6E \\
    % \midrule
    \midrule
  \sc  P-E & 58.4\textsubscript{2.6} & $-$ & $-$ \\
    % \sc E-P & \bf 64.4$\pm$2.9 & $-$ & $-$ \\
  \cmidrule(r){1-1}
  \sc  P-E+ExplCal &  $-$ & {\bf 62.2}\textsubscript{2.8} & {\bf 62.6}\textsubscript{3.1}\\
    \bottomrule
    \end{tabular}
\end{table}

We also experiment with calibrating predictions obtained using {\sc P-E} on \hotpot{}, following the same setup as described in Section~\ref{sec:calib_hotpot}.

As in Table~\ref{tab:app_hotpot}, using explanations is also effective for calibrating the predictions obtained using {\sc P-E}. {\sc P-E+ExplCal} successfully improves the AUC of {\sc P-E} from 58.4 to 62.6, outperforming  {\sc Few-Shot} and {\sc Few-Shot (NN)}.

\section{Additional Output Examples}
\label{sec:app_add_exs}

\begin{figure}[h]
    \centering
    \footnotesize
    \begin{tabular}{l|l}
    \toprule
    \multicolumn{2}{c}{\sc \synth{}: E-P} \\
    \midrule
  \multirow{8}{*}{\STAB{\rotatebox[origin=c]{90}{\bf Nonfactual}}}
& Jason helps Christopher. Kevin helps Stephanie. James punches Jonathan. Daniel is a typist. \textcolor{red}{Jonathan} \\
&\textcolor{red}{is a tailor}. Christopher is a typist. Richard punches Daniel. Stephanie is a tailor. \\
& Q:  Who punches a typist?\\
& A: Because \textcolor{red}{Jonathan is a typist} and James punches Jonathan, the answer is \textcolor{red}{James}.\\
\cmidrule{2-2}
& Matthew blames Tiffany. Lisa is a chef. Christopher helps Kelly. Angela helps Jessica. \\ 
& Rachel blames Lisa. Jessica is a farmer. \textcolor{red}{Kelly is a chef}. Tiffany is a farmer. \\
& Q: Who helps a farmer? \\
& A: Because \textcolor{red}{Kelly is a farmer} and Christopher helps Kelly, the answer is \textcolor{red}{Christopher}. \\
\midrule

\multicolumn{2}{c}{\sc \synth{}: P-E} \\
    \midrule
    \multirow{8}{*}{\STAB{\rotatebox[origin=c]{90}{\bf Nonfactual}}} 
& Richard is a governor. Megan supervises Adam. Tiffany calls Nicholas. \textcolor{red}{Nicholas is a musician}. Elizabeth \\ &
is a musician. Matthew calls Richard. Robert supervises Elizabeth. Adam is a governor. \\
&Q: Who calls a governor?\\
&A:  \textcolor{red}{Tiffany}, because \textcolor{red}{Nicholas is a governor} and Tiffany calls Nicholas.\\
\cmidrule{2-2}
& Anthony believes Justin. John is an engineer. Eric answers Kevin. \textcolor{red}{Amy is an engineer}. Justin \\
& is a pilot. Sarah answers Amy. Kevin is a pilot. Brandon believes John.\\
&Q: Who answers a pilot?\\
&A: \textcolor{red}{Sarah}, because \textcolor{red}{Amy is a pilot} and Sarah answers Amy. \\
    \bottomrule
    \end{tabular}

    \caption{Examples of failure cases on \synth{}. \gpt{} almost always generates nonfactual explanations when giving an incorrect answer, using either {\sc P-E} or {\sc E-P}. }
    \label{fig:app_synth_error}
\end{figure}

\begin{figure}[h]
    \centering
    \footnotesize
    \begin{tabular}{l|l}
    \toprule
    \multicolumn{2}{c}{\sc \hotpot{}: P-E} \\
    \midrule
  \multirow{8}{*}{\STAB{\rotatebox[origin=c]{90}{\bf Nonfactual}}} &
"Crestfallen" is a track on "Adore". The single's artwork is by Yelena Yemchuk. \\
 &Johnny McDaid is a Croatian professional photographer, painter and film director. \\ 
&Arnljot Berg is a French professional photographer, painter and film director. \\
&\textcolor{red}{Yelena Yemchuk is a Ukrainian professional photographer, painter and film director.} \\
&Q: Crestfallen's artwork is done by a photographer of which nationality? \\
&A: \textcolor{red}{Croatian}. The reasons are as follows. First, Crestfallen's artwork is done by Yelena Yemchuk. \\
& Second, \textcolor{red}{Yelena Yemchuk is Croatian photographer}.\\
    % \cmidrule{2-2}
    \midrule
    \multirow{7}{*}{\STAB{\rotatebox[origin=c]{90}{\bf Inconsistent}}} & The Pequannoc Spillway is a professional basketball team based in Ellesmere city, United Kingdom.\\
    & The Silver Lake  is a professional basketball team based in Ellesmere harbour, United Kingdom.\\
    & The Cheshire Phoenix is a professional basketball team based in Ellesmere Port, United Kingdom. \\
    &Mike DiNunno (born January 29, 1990) is an American-Italian professional basketball player for the \\ & Cheshire Phoenix, playing the point guard position.\\
    & Q: Where is the basketball team that Mike DiNunno plays for based? \\
    & A: \textcolor{red}{Cheshire Phoenix}. The reason is as follows. First, Mike DiNunno plays for the Cheshire Phoenix. \\ & Second,  \textcolor{red}{the Cheshire Phoenix is based in Ellesmere Port}. \\
    \bottomrule

    \end{tabular}

    \caption{Examples of unreliable explanations generated for \hotpot{} using {\sc P-E}.}
    \label{fig:app_hotpot_error}
\end{figure}

\begin{figure}[h]
    \centering
    \footnotesize
    \begin{tabular}{l|l}
    \toprule
    \multicolumn{2}{c}{\sc \esnli{}: E-P} \\
    \midrule
  \multirow{6}{*}{\STAB{\rotatebox[origin=c]{90}{GT: {\bf True}}}}
  
  & A female is looking through a microscope. \\
& Q: A lady is observing something True, False, or Neither? \\
& A: True, because if you're looking through a microscope you are observing something. \\
\cmidrule{2-2} 
% & A person powerwashing stairs leading down to a pool. \\
% & Q: a person washes stairs True, False, or Neither? \\
% & A: True, because if you powerwash, you are washing. \\
% \cmidrule{2-2}
&A person wearing sunglasses, a visor, and a British flag is carrying 6 Heineken bottles. \\
&Q: The person has beer bottles and has a england flag and glasses. True, False, or Neither? \\
&A: \textcolor{red}{False} , because the person could be carrying any number of items and not just six Heineken bottles. \\
\midrule

\multirow{6}{*}{\STAB{\rotatebox[origin=c]{90}{GT: {\bf False}}}}

& A woman and a man are sword fighting. \\
& Q: Two people have engaged in a sword fight and both are dead. True, False, or Neither? \\
& A: False, because it is impossible for two people to sword fight and both be dead. \\
\cmidrule{2-2}
& Two construction workers in a truck having a conversation. \\
& Q: Two people are outside a truck. True, False, or Neither? \\
& A: \textcolor{red}{True}, because two people are outside a truck. \\
\midrule
\multirow{6}{*}{\STAB{\rotatebox[origin=c]{90}{GT: {\bf Neither}}}}
& A dog running on grass \\
& Q: A dog is playing outside. True, False, or Neither?\\
& A: \textcolor{red}{True}, because a dog playing outside implies that the dog is playing outside.\\
\cmidrule{2-2}
& A silhouette at the bottom of an escalator. \\
& Q: The silhoutte is creeping out the children. True, False, or Neither? \\
& A: \textcolor{red}{False}, because the silhouette is not necessarily creeping out the children. \\
\bottomrule
    \end{tabular}

    \caption{The completions generated for \esnli{} examples with different ground truth labels (GT)  using {\sc E-P}. \gpt{} sometimes ignores the information from premises when explaining its predictions (examples in the bottom section).}
    \label{fig:app_error}
\end{figure}

\newpage
\section{Details of Automatically Assessing Consistency and Factuality on \synth{}}
Our questions follow the template \texttt{\small Who $V_1$ $P_1$?}. Our generated explanations follow the template  \texttt{\small $N_1$ is $P_2$ and $N_2$ $V_2$ $N_3$}. Our answers are always a name, e.g., \texttt{\small $N_4$}. Because large language models almost always produce well-formed explanations, we can match the output against these patterns and extract variables $V_1$, $P_1$, etc. using simple regular expressions.

We say that an explanation is consistent if and only if the following conditions are satisfied: (1) \texttt{\small $N_2$ = $N_4$} and \texttt{\small $N_1$ = $N_3$}. (2)  \texttt{\small $P_2$ = $P_1$} and \texttt{\small $V_2$ = $V_1$}. These rules ensure the explanation matches the intent of the question and entails the answer at the same time.

We say an explanation is factual if and only if both \texttt{\small $N_1$ is $P_2$} and \texttt{\small  $N_2$ $V_2$ $N_3$} appear exactly in the context.

\section{Results of Using Explanations in an Alternative Style on \synth{}}
\label{sec:alternative}

\begin{table}[h]
  \caption{Performance of text-davinci-001 of using explanations in an alternative style on \synth{}. }
  \label{tab:altprompting}
  \centering
  \begin{tabular}{clc}
    \toprule
    % \multicolumn{2}{c}{Part}                   \\
    % \cmidrule(r){1-2}
    && \synth{} \\
    \midrule
    \multirow{3}{*}{GPT-3} &
    \sc Few-Shot & \bf 49.5$\pm$0.6 \\
    \cmidrule{2-3}
    & \sc E-P (Alternative) & 48.0$\pm$2.6 \\
    & \sc P-E (Alternative) & 49.5$\pm$1.7 \\
    \midrule
    \multirow{3}{*}{InstructGPT} &
    \sc Few-Shot & \bf 54.8$\pm$2.5 \\
    \cmidrule{2-3}
    & \sc E-P (Alternative) & 50.6$\pm$1.6 \\
    & \sc P-E (Alternative) & 53.3$\pm$1.6 \\
    \midrule
    \multirow{3}{*}{text-davinci-002} &
    \sc Few-Shot & 72.0$\pm$1.4 \\
    \cmidrule{2-3}
    & \sc E-P (Alternative) & 75.3$\pm$2.2 \\
    & \sc P-E (Alternative) & \bf 80.5$\pm$2.4 \\
    \bottomrule
  \end{tabular}
\end{table}

\begin{table}
    \caption{Reliability of explanations in an alternative style.}
  \label{tab:altreliability}
  \centering

    \begin{tabular}{clccc|cc}
    \toprule
    % \multicolumn{2}{c}{Part}                   \\
    % \cmidrule(r){1-2}
  && Acc & Fac & Con &  Acc=Fac & Acc=Con\\
    \midrule
      \multirow{2}{*}{davinci}
    & {\sc Synth} ({\sc Alternative; E-P})&  48.4 & 53.6  & 98.4  & \bf 94.8 & 48.4\\
    & {\sc Synth} ({\sc Alternative; P-E})&  51.6 & 53.2  & 100.  & \bf 98.4 & 51.6 \\  
    \midrule
    \multirow{2}{*}{text-davinci-001}
    & {\sc Synth} ({\sc Alternative; E-P})&  50.8 & 53.6  & 97.6  & \bf 97.2 & 53.2\\
    & {\sc Synth} ({\sc Alternative; P-E})&  52.8 & 52.8  & 98.4  & \bf 98.4 & 54.8 \\    
    \midrule
    \multirow{2}{*}{text-davinci-002}
    & {\sc Synth} ({\sc Alternative; E-P})&  75.2 & 79.6  & 100.  & \bf 95.6 & 75.2 \\
    & {\sc Synth} ({\sc Alternative; P-E})&  82.8 & 86.0  & 100.  & \bf 96.8 & 82.8 \\    
    \bottomrule
  \end{tabular}
  
\end{table}

We also experimented with using an alternative style of explanations for \synth{}, where we reversed the order of the two sentences in the explanations shown in Table~\ref{fig:data_exs}. These explanations follow the format: \texttt{\small A [verb] B and B is [profession].} (instead of \texttt{\small B is [profession] and A [verb] B.}) By changing the order in which the sentences are extracted, we might expect that E-P can more easily follow the reasoning chain.

We show the performance of using reversed explanations in Table~\ref{tab:altprompting} and the reliability in Table~\ref{tab:altreliability}. In general, this alternative style of explanations yields inferior performance compared to the original style (Table~\ref{tab:promptingLLMs}).
Using explanations leads to no improvements on GPT-3, and \davone{}. {\sc P-E} is consistently better than {\sc E-P} across GPT-3, \davone{}, and \davtwo{}. 

Furthermore, using such a reversed style, language models almost always generate consistent explanations when being prompted in either {\sc E-P} or {\sc P-E} paradigm. The factuality almost always indicates the correctness of predictions.

We believe these two prompts cover the most natural explanation styles for this problem. While small format changes or modifications to the general QA prompt format are also possible, we observed these to have minor impacts on the results (as we see in Appendix~\ref{sec:stepbystep}).

\section{Results of Adding ``Step by Step'' Trigger in Prompts}
\label{sec:stepbystep}
We test whether including a trigger for multi-step reasoning can help LLMs better learn from explanations in the prompt for multi-step reasoning. Following \citet{zerocot}, we prepend ``Let's think step by step.'' to the exemplar explanations used in the {\sc E-P} paradigm. For this experiment, we only test on \synth{} and \hotpot{}, which involve multi-step reasoning. We do not experiment with text-davinci-002, which has already achieved substantial performance improvement from using explanations, and we omit OPT because its performance is too low.

As shown in Table~\ref{tab:addtrigger}, adding triggers in the prompts does not lead to statistically significantly improvements in E-P for GPT-3 and InstructGPT. In fact, it typically causes a performance degradation. 

\begin{table}
  \caption{Results of adding ``let's think step by step'' trigger in prompts.}
  \label{tab:addtrigger}
  \centering
  % \scriptsize
%   \foo
  \begin{tabular}{clcc}
    \toprule
    % \multicolumn{2}{c}{Part}                   \\
    % \cmidrule(r){1-2}
    
     && \synth{} & \hotpot{} \\

    \midrule
     \multirow{3}{*}{davinci} 
    & \sc Few-Shot & {\bf 49.5}\textsubscript{0.6}  & 49.1\textsubscript{6.2}\\
    \cmidrule{2-4}
    & \sc E-P & {47.1}\textsubscript{2.8} & {\bf 54.1}\textsubscript{4.1}\\
    & \sc E-P + Trigger & {48.6}\textsubscript{2.6} & 50.1\textsubscript{5.2}\\
    
    \midrule
    \multirow{3}{*}{text-davinci-001} 
    & \sc Few-Shot & {54.8}\textsubscript{2.5}  & 53.2\textsubscript{2.3}\\
    \cmidrule{2-4}
    & \sc E-P & {\bf 58.5}\textsubscript{2.1} & {\bf 58.2}\textsubscript{4.1}\\
    & \sc E-P + Trigger & 58.0\textsubscript{3.4} & 58.0\textsubscript{6.2}\\
    \bottomrule
  \end{tabular}
\end{table}

\section{Information about Cost of Running Experiments}

The cost of our experiments, described as follows, is estimated based on using the GPT-3 API with the largest models available (davinci, text-davinci-001, and text-davinci-002) as of August 2022 (\$0.06 per 1,000 tokens). The setting in Table~\ref{tab:promptingLLMs} uses 250 examples for each result, with roughly 1400 tokens per example using the {\sc Few-Shot} paradigm and 2000 tokens per example using the {\sc E-P} or {\sc E-P} paradigm. The cost of evaluating {\sc Few-Shot}, {\sc P-E}, and {\sc E-P} for 5 trials on a single dataset is roughly \$105, \$150, and \$150, respectively. The total price for reproducing results on three datasets as in Table~\ref{tab:promptingLLMs} using a single language model is roughly \$1200.

We subsample 250-example sets to reduce cost rather than running on full datasets. Based on the significance tests in this paper and the reported confidence intervals, this size dataset is sufficient to distinguish between the performance of different approaches.

\end{document}